\pdfoutput=1

\documentclass[11pt]{article}

\usepackage{emnlp2021}

\usepackage{times}
\usepackage{latexsym}

\usepackage{microtype}
\usepackage{amsmath,amsfonts,amssymb}
\usepackage{multirow}
\usepackage{xspace}
\usepackage{subcaption}

\usepackage{appendix}
\usepackage{etoolbox}
\usepackage{stfloats}
\usepackage{graphicx}
\usepackage[ruled,vlined]{algorithm2e}
\usepackage{enumitem}
\usepackage{algpseudocode}
\usepackage{floatrow}
\usepackage{bm}
\usepackage{bbm}
\newfloatcommand{capbtabbox}{table}[][\FBwidth]

\usepackage{blindtext}
 
 \usepackage{wrapfig}
\usepackage{booktabs}
\usepackage{xspace}
\usepackage{placeins}

\usepackage[T1]{fontenc}

\usepackage[utf8]{inputenc}

\usepackage{microtype}

\newcommand{\system}[1]{\textsc{#1}\xspace}

\newcommand{\blfs}{\texttt{DLFS}\xspace}
\newcommand{\aemr}{\texttt{TR}\xspace}
\newcommand{\awu}{\texttt{FSCL}\xspace}
\newcommand{\TR}{\texttt{TR}\xspace}
\newcommand{\tr}{\system{Total Recall}}

\newcommand{\dlfs}{\texttt{DLFS}\xspace}
\newcommand{\dar}{\texttt{DAR}\xspace}

\newcommand{\data}[1]{\textsc{#1}\xspace}

\newcommand{\overnight}{\data{Overnight}}
\newcommand{\nlmapc}{\data{NLMap(city)}}
\newcommand{\nlmapq}{\data{NLMap(qt)}}

\newcommand{\seqseq}{\system{Seq2Seq}}

\newcommand{\blfst}{\system{DLFS}}
\newcommand{\aemrt}{\system{TR}}

\newcommand{\finetune}{\system{Fine-tune}}
\newcommand{\hatt}{\system{HAT}}
\newcommand{\ewc}{\system{EWC}}

\newcommand{\gem}{\system{GEM}}
\newcommand{\emar}{\system{EMAR}}
\newcommand{\arper}{\system{ARPER}}
\newcommand{\emr}{\system{EMR}}
\newcommand{\proto}{\system{ProtoParser}}
\newcommand{\oracle}{\system{Oracle}}

\newcommand{\gss}{\system{GSS}}
\newcommand{\balance}{\system{Balance}}
\newcommand{\fss}{\system{FSS}}
\newcommand{\prior}{\system{Prior}}
\newcommand{\lfs}{\system{LFS}}
\newcommand{\random}{\system{Random}}

\usepackage{amsthm,amsmath,amsfonts,bm,xspace}
\usepackage{color}

\newcommand{\comment}[1]{}

\def\eqref#1{(\ref{#1})}

\def\1{\bm{1}}

\def\va{{\bm{a}}}

\def\vx{{\bm{x}}}
\def\vy{{\bm{y}}}

\DeclareMathAlphabet{\mathsfit}{\encodingdefault}{\sfdefault}{m}{sl}
\SetMathAlphabet{\mathsfit}{bold}{\encodingdefault}{\sfdefault}{bx}{n}

\newcommand{\R}{\mathbb{R}}

\newcommand{\softmax}{\mathrm{softmax}}

\newcommand{\todo}[1]{\textcolor{red}{TODO}}

\title{Total Recall: a Customized Continual Learning Method for Neural Semantic Parsers}

\author{Zhuang Li, Lizhen Qu\thanks{corresponding author}, Gholamreza Haffari\\
Faculty of Information Technology\\
  Monash University \\
  {\tt firstname.lastname@monash.edu}}

\begin{document}

\maketitle
\abovedisplayskip=0.2pt
\abovedisplayshortskip=0.2pt
\belowdisplayskip=0.2pt
\belowdisplayshortskip=0.2pt

\begin{abstract}
This paper investigates continual learning for semantic parsing. In this setting, a neural semantic parser learns tasks sequentially without accessing full training data from previous tasks. Direct application of the SOTA continual learning algorithms to this problem fails to achieve comparable performance with re-training models with all seen tasks, because they have not considered the special properties of structured outputs, yielded by semantic parsers. Therefore, we propose \tr, a continual learning method designed for neural semantic parsers from two aspects: i) a sampling method for memory replay that diversifies logical form templates and balances distributions of parse actions in a memory; 
ii) a two-stage training method that significantly improves generalization capability of the parsers across tasks. We conduct extensive experiments to study the research problems involved in continual semantic parsing, and demonstrate that a neural semantic parser trained with \tr achieves superior performance than the one trained directly with the SOTA continual learning algorithms, and achieve a 3-6 times speedup compared to retraining from scratch. Code and datasets are available at: \url{https://github.com/zhuang-li/cl_nsp}. 
\end{abstract}

\section{Introduction}
In the recent market research report published by MarketsandMarkets~\cite{MarketsandMarkets}, it is estimated that the smart speaker market is expected to grow from USD 7.1 billion in 2020 to USD 15.6 billion by 2025. Commercial smart speakers, such as Alexa and Google assistant, often need to translate users' commands and questions into actions. Therefore, semantic parsers are widely adopted in dialogue systems to map natural language (NL) utterances to executable programs or logical forms (LFs)~\cite{damonte2019practicalSemanticParsing,rongali2020seq2seqSemanticParsing}. Due to the increasing popularity of such speakers, software developers have implemented a large volume of skills for them and the number of new skills grow quickly every year. For example, as of 2020, the number of Alexa skills exceeds 100,000 and 24 new skills are introduced per day in 2020~\cite{voicebot}. Although machine learning-based semantic parsers achieve the state-of-the-art performance, they face the following challenges due to the fast growing number of tasks. 

Given new tasks, one common practice is to re-train the parser from scratch on the training data of all seen tasks. However, it is both economically and computationally expensive to re-train semantic parsers because of a fast-growing number of new tasks~\cite{lialin2020clNeuralSemanticParsing}. To achieve its optimal performance, training a deep model on all 8 tasks of NLMap~\cite{lawrence2018improving} takes approximately 14 times longer than training the same model on one of those tasks. 
In practice, the cost of repeated re-training for a commercial smart speaker is much higher, e.g. Alexa needs to cope with the number of tasks which is over 10,000 times more than the one in NLMap\footnote{A rough estimation: re-training of our semantic parser for 100,000 tasks will take more than 138 days (2 mins of training time per NLMap task$\times$100,000) on a single-GPU machine.}. In contrast, \textit{continual learning} provides an alternative cost-effective training paradigm, which learns tasks sequentially without accessing full training data from the previous tasks, such that the computational resources are utilized only for new tasks. 

Privacy leakage has gradually become a major concern in many Artificial Intelligence (AI) applications. As most computing environments are not 100\% safe, it is not desirable to always keep a copy of the training data including identifiable personal information. Thus, it is almost not feasible to assume that complete training data of all known tasks is always available for re-training a semantic parser~\cite{irfan2021lifelong}. For the semantic parser of a privacy-sensitive AI system, e.g. personalized social robot, \textit{continual learning} provides a solution to maintain the knowledge of all learned tasks when the complete training data of those data is not available anymore due to security reasons. 
%


A major challenge of continual learning lies in \textit{catastrophic forgetting} that the (deep) models easily \textit{forget} the knowledge learned in the previous tasks when they learn new tasks~\cite{french1991using,mi2020continual}. Another challenge is to learn what kind of knowledge the tasks share in common and support fast adaptation of models for new tasks. Methods are developed to mitigate catastrophic forgetting~\cite{lopez2017gradient,han2020continual} and facilitate forward knowledge transfer~\cite{li2017learning}. Instead of directly measuring speedup of training, those methods assume that there is a small fixed-size memory available for storing training examples or parameters from the previous tasks. The memory limits the size of training data thus proportionally reduces training time. However, we empirically found that direct application of those methods to neural semantic parsers leads to a significant drop of test performance on benchmark datasets, in comparison to re-training them with all available tasks each time. 

In this work, we investigated the applicability of existing continual learning methods to semantic parsing in-depth, and we have found that most methods have not considered the special properties of structured outputs, which distinct semantic parsing from the multi-class classification problem. Therefore, we propose \tr (\TR), a continual learning method that is especially designed to address the semantic parsing specific problems from two perspectives. First, we customize the sampling algorithm for memory replay, which stores a small sample of examples from each previous task when continually learning new tasks. The corresponding sampling algorithm, called \textbf{D}iversified \textbf{L}ogical \textbf{F}orm \textbf{S}election (\dlfs), diversifies LF templates and maximizes the entropy of the parse action distribution in a memory. 
Second, motivated by findings from cognitive neuroscience~\cite{goyal2020inductiveBias}, we facilitate knowledge transfer between tasks by proposing a two-stage training procedure, called \textbf{F}ast \textbf{S}low \textbf{C}ontinual \textbf{L}earning (\awu). It updates only unseen action embeddings in the fast-learning stage and updates all model parameters in the follow-up stage. As a result, it significantly improves generalization capability of parsing models.

Our key contributions are as follows:
\begin{itemize}
    \item We conduct the \textit{first} in-depth empirical study of the problems encountered by neural semantic parsers to learn a \textit{sequence} of tasks continually in various settings. The most related work~\cite{lialin2020clNeuralSemanticParsing} only investigated incremental learning between two semantic parsing tasks.
    \item We propose \dlfs, a sampling algorithm for memory replay that is customized for semantic parsing. As a result, it improves the best sampling methods of memory replay by 2-11\% on Overnight~\cite{wang2015overnight}.
    \item We propose a two-stage training algorithm, coined \awu, that improves the test performance of parsers across tasks by 5-13\% in comparison with using only Adam~\cite{kingma2014adam}.
    \item In our extensive experiments, we investigate applicability of the SOTA continual learning methods to semantic parsing with \textit{three} different task definitions, and show that \TR outperforms the competitive baselines by 4-9\% and achieves a speedup by 3-6 times compared to training from scratch. 
\end{itemize}


\section{Related Work}
\paragraph{Semantic Parsing} The recent surveys~\cite{Kamath2018semanticParsingSurvey,zhu2019survey,li2020context} cover an ample of work in semantic parsing. Most current work employ a sequence-to-sequence architecture~\cite{sutskever2014sequence} to map an utterance into a structured meaning representations, such as LFs, SQL, and abstract meaning representation~\cite{banarescu2013AMR}. %
The output sequences are either linearized LFs~\cite{dong2016language,dong2018coarse,cao2019semantic} or sequences of parse actions~\cite{chen2018sequence, cheng2019executableParser, lin2019grammarText2SQL, zhang2019broadTransduction,yin2018tranx,chen2018sequence,guo2019irnet,wang2019rat,li2021few}. 
There are also work~\cite{guo2019irnet,wang2019rat,li2021few} exploring semantic parsing with unseen database schemas or actions. Feedback semantic parsing interactively collects data from user feedbacks as continuous data streams but does not address the problem of catastrophic forgetting or improve forward transfer~\cite{iyer2017learning,yao2019model,labutov2019learning}. 

\paragraph{Continual Learning}

The continual learning methods can be coarsely categorized into i) regularization-based methods~\cite{kirkpatrick2017overcoming,zenke2017continual,ritter2018online,li2017learning,zhao2020maintaining,schwarz2018progress} which either applies knowledge distillation~\cite{hinton2015distilling} to penalize the loss updates or regularizes parameters which are crucial to the old tasks; ii) dynamic architecture methods~\cite{mallya2018packnet,serra2018overcoming,maltoni2019continuous,houlsby2019parameter,wang2020k,pfeiffer2020adapterfusion,rusu2016progressive} which dynamically alter the structures of models to reduce the catastrophic forgetting; iii) memory-based methods~\cite{lopez2017gradient,wang2019sentence,han2020continual,aljundi2019gradient,chrysakis2020online,kim2020imbalanced} which stores historical instances and continually learn them along with instances in new tasks. There are also hybrid methods~\cite{mi2020continual,liu2020mnemonics,rebuffi2017icarl} which integrate more than one type of such methods. 

In natural language processing (NLP), continual learning is applied to tasks such as relation extraction~\cite{wang2019sentence,han2020continual}, natural language generation~\cite{mi2020continual}, language modelling~\cite{sun2019lamol} and the pre-trained language models adapting to multiple NLP tasks~\cite{wang2020k,pfeiffer2020adapterfusion}. To the best of our knowledge, \cite{lialin2020clNeuralSemanticParsing} is the only work studying catastrophic forgetting for semantic parsing. However, they consider learning between \textit{only} two tasks, have not proposed new methods, and also have not evaluated recently proposed continual learning methods. In contrast, we propose two novel continual learning methods customized for semantic parsing and compare them with the strong and recently proposed continual learning methods that are not applied to semantic parsing before.


\section{Base Parser}
A semantic parser learns a mapping $\pi_{\theta} : \mathcal{X} \rightarrow \mathcal{Y}$ to convert an natural language (NL) utterance $\vx \in \mathcal{X}$ into its corresponding logical form (LF) $\vy \in \mathcal{Y}$. Most SOTA neural semantic parsers formulate this task as translating a word sequence into an output sequence, whereby an output sequence is either a sequence of LF tokens or a sequence of parse actions that construct an LF. For a fair comparison between different continual learning algorithms, we adopt the same base model for them, as commonly used in prior works~\cite{lopez2017gradient,wang2019sentence,han2020continual}.

Similar to~\cite{shin2019programSynthesisIdioms,iyer2019learningIdiomsSP}, the base parser converts the utterance $\vx$ into a sequence of actions $\va = \{a_1, ..., a_t\}$. 
As an LF can be equivalently parsed into an abstract syntax tree (AST),
the actions $\va$ sequentially construct an AST deterministically 
in the depth-first order, wherein each action $a_t$ at time step $t$ either i) expands an intermediate node according to the production rules from a grammar, 
or ii) generates a leaf node. As in ~\cite{shin2019programSynthesisIdioms}, the idioms (frequently occurred AST fragments) are collapsed into single units. 
The AST is further mapped back to the target LF. 

The parser employs the attention-based sequence-to-sequence (\seqseq) architecture~\cite{luong2015effective} for estimating action probabilities.

\vspace{-3mm}
\begin{small}
\begin{equation}
    P(\va | \vx ) = \prod_{t = 1}^{|\va|} P(a_t | \va_{<t},\vx)
\end{equation}
\end{small}

\paragraph{Encoder.} The encoder in \seqseq is a standard bidirectional Long Short-term Memory (LSTM) network~\cite{hochreiter1997long}, which encodes an utterance $\vx$ into a sequence of contextual word representations.


\paragraph{Decoder.} 
The decoder applies an LSTM to generate action sequences. At time $t$, the decoder produces an action representation $\mathbf{s}_t$, which is yielded by concatenating the hidden representation $\mathbf{h}_t$ produced by the LSTM and the context vector $\mathbf{o}_t$ produced by the soft attention~\cite{luong2015effective}. 




We maintain an embedding for each action in the embedding table. The probability of an action $a_t$ is estimated by:

\vspace{-3mm}
\begin{small}
\begin{equation}
\label{eq:action_prob}
 P(a_t | \va_{<t},\vx) =  \frac{\exp(\mathbf{c}_{a_t}^{\intercal} \mathbf{s}_t)}{\sum_{a' \in \mathcal{A}_t}\exp(\mathbf{c}_{a'}^{\intercal} \mathbf{s}_t)}
\end{equation}
\end{small}
where $\mathcal{A}_t$ is the set of applicable actions at time $t$, and $\mathbf{c}_{a}$ is the embedding of the action $a_{t}$, which is referred to as \textit{action embedding} in the following.

\section{Continual Semantic Parsing}
\paragraph{Problem Formulation.}
We consider a widely adopted continual learning setting that a parser $\pi_{\theta}$ is trained continually on a sequence of $K$ distinct tasks \{$\mathcal{T}^{(1)}$, $\mathcal{T}^{(2)}$,...,$\mathcal{T}^{(K)}$\}~\cite{lopez2017gradient,han2020continual}. 
In both training and testing, we know which task an example belongs to. As the definition of tasks is application specific and parallel data of semantic parsing is often created by domain experts, it is easy to identify the task of an example in practice. We further assume that there is a fixed-size memory $\mathcal{M}_{k}$ associated with each task $\mathcal{T}^{(k)}$ for e.g. storing a small amount of replay instances, as adopted in~\cite{rebuffi2017icarl,wang2019sentence}. This setting is practical for personalized conversational agents because it is difficult for them to re-collect past information except reusing the ones in the memories.

\subsection{Challenges}
\label{sec:challenges}
We demonstrate catastrophic forgetting in continual semantic parsing by training the base parser sequentially with each task from the \overnight corpus~\cite{wang2015overnight} and report the test accuracy of exactly matched LFs of all seen tasks combined (More evaluation details are in Sec. \ref{sec:exp}).

\begin{figure}[ht]
           \vspace{-2mm}
    \centering
    \includegraphics[width=0.9\textwidth]{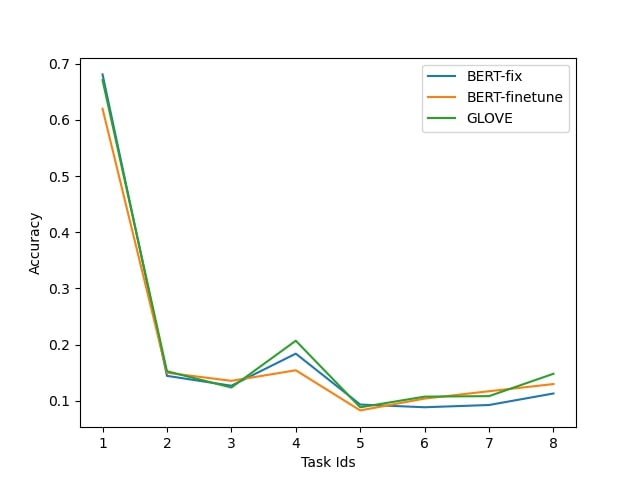}
    \caption{The accuracy on \overnight after the parser being trained on each task. The parser uses GLOVE~\cite{pennington2014glove}, BERT~\cite{devlin2019bert} with parameters updated (BERT-finetune) and fixed (BERT-fix) as the input embedding layer. \vspace{-4mm}}
    \label{fig:all_embeddings_acc_overnight}
        \vspace{-2mm}
\end{figure}
Fig.~\ref{fig:all_embeddings_acc_overnight} shows the performance of continually training the base parser with BERT~\cite{devlin2019bert} (with and without fine-tuning BERT parameters) and GlOVE respectively by using the standard cross entropy loss. The accuracy on the combined test set drops dramatically after learning the \textit{second} task. The training on the initial task appears to be crucial on mitigating catastrophic forgetting. The BERT-based parser with/without fine-tuning obtains no improvement over the one using GLOVE. The forgetting with BERT is even more serious compared with using GLOVE. The same phenomenon is also observed in~\cite{arora2019forgettingBaldwin} that the models with pre-trained language models obtain inferior performance than LSTM or CNN, when fine-tuning incrementally on each task. They conjecture that it is more difficult for models with large capacity to mitigate catastrophic forgetting. 

\begin{figure}[ht]
           \vspace{-2mm}
    \centering
    \includegraphics[width=0.9\textwidth]{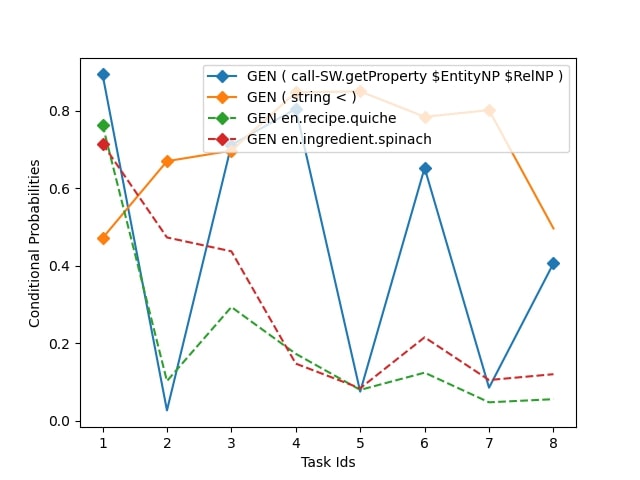}
    \caption{The average conditional probabilities $P(a_t | \va_{<t},\vx)$ of the representative cross-task (solid) and task-specific (dash) actions till the seen tasks on \overnight after learning on each task sequentially. The boxes at $i$th task indicate the actions from the initial task also exist in the $i$th task. \vspace{-4mm}}
   \label{fig:agg_acc_overnight}
           \vspace{-2mm}
\end{figure}
We further investigate which parse actions are easy to forget. To measure the degree of forgetness of an action, after training the parser in the first task, we average the probabilities $P(a_t | \va_{<t},\vx)$ produced by the parser on the training set of the first task. We recompute the same quantity after learning each task sequentially and plot the measures. Fig. \ref{fig:agg_acc_overnight} depicts the top two and the bottom two actions are easiest to forget on average. Both top two actions appear only in the first task, thus it is difficult for the parser to remember them after learning new tasks. In contrast, cross-task actions, such as \textit{GEN ( string < )}, may even obtain improved performance after learning on the last task. Thus, it indicates the importance of differentiating between task-specific actions and cross-task actions when designing novel continual learning algorithms.



\subsection{\tr}
To save training time for each new task, we cannot use all training data from previous tasks, thus we introduce a designated sampling method in the sequel to fill memories with the examples most likely mitigating catastrophic forgetting. We also present the two-stage training algorithm \awu to facilitate knowledge transfer between tasks.

\paragraph{Sampling Method.}
\dlfs improves Episodic Memory Replay (\emr)~\cite{wang2019sentence,chaudhry2019tiny} by proposing a designated sampling method for continual semantic parsing. \emr utilizes a memory module $\mathcal{M}_{k} = \{(\vx_1^{(k)},\vy_1^{(k)}),...,(\vx_M^{(k)},\vy_M^{(k)})\}$ to store a few examples sampled from the training set of task $\mathcal{T}^{(k)}$, where $(\vx_m^{(k)},\vy_m^{(k)}) \in \mathcal{D}_{train}^{(k)}$ and $M$ is the size of the memory. The training loss of \emr takes the following form:
\begin{equation}
    \mathcal{L}_{\emr} = \mathcal{L}_{\mathcal{D}_{train}^{(k)}} + \sum_{i=1}^{k-1}\mathcal{L}_{\mathcal{M}_{i}}
\end{equation}
where $\mathcal{L}_{\mathcal{D}_{train}^{(k)}}$ and $\mathcal{L}_{\mathcal{M}_{i}}$ denotes the loss on the training data of current task $\mathcal{T}^{(k)}$ and the memory data from task $\{\mathcal{T}\}^{(k-1)}_{i=1}$, respectively. The training methods for memory replay often adopt a subroutine called \textit{replay training} to train models on instances in the memory. Furthermore, prior works~\cite{aljundi2019gradient,wang2019sentence,han2020continual,mi2020continual,chrysakis2020online,kim2020imbalanced} discovered that storing a small amount of \textit{diversified} and \textit{long-tailed} examples is helpful in tackling catastrophic forgetting for memory-based methods. Semantic parsing is a structured prediction problem. We observe that semantic parsing datasets are highly imbalanced w.r.t. LF structures. Some instances with similar LF structures occupy a large fraction of the training set. Therefore, we presume storing the diversified instances in terms of the corresponding LF structures would alleviate the problem of catastrophic forgetting in continual semantic parsing. 

To sample instances with the diversified LF structures, our method \dlfs partitions the LFs in $\mathcal{D}_{train}$ into $M$ clusters, followed by selecting representative instances from each cluster to maximize entropy of actions in a memory. To characterize differences in structures, we first compute similarities between LFs by $sim(\vy_i,\vy_j) = (Smatch(\vy_i,\vy_j) + Smatch(\vy_j,\vy_i))/2$, where \textit{Smatch}~\cite{cai2013smatch} is a asymmetrical similarity score between two LFs yielded by calculating the overlapping percentage of their triples. Then we run a flat clustering algorithm using the distance function $1- sim(\vy_i,\vy_j)$ and the number of clusters is the same as the size of a memory. We choose \textit{K-medoids}~\cite{park2009simple} in this work for easy interpretation of clustering results.

We formulate the problem of balancing action distribution and diversifying LF structures as the following constrained optimization problem. In particular, it i) aims to balance the actions of stored instances in the memory module $\mathcal{M}$ by increasing the entropy of the action distribution, and ii) requires that each instance $m$ in $\mathcal{M}$ belongs to a different cluster $c_j$. Let the function $c(m)$ return the cluster id of an instance in a memory $\mathcal{M}$ and $m_i$ denote its $i$th entry, we have
\begin{align}\nonumber
\label{eq:sample}
 \max_{\mathcal{M}}& -\sum_{a_{i} \in \mathcal{A}} p_{\mathcal{M}}(a_{i})\log p_{\mathcal{M}}(a_{i})\\
    s.t. & \forall m_i, m_j \in \mathcal{M}, c(m_i) \neq c(m_j)
\end{align}
\noindent where $p_{\mathcal{M}}(a_{i})=\frac{n_{i}}{\sum_{{a_j} \in \mathcal{A}} n_{j}}$, with $n_i$ being the frequency of action $a_{i}$ in $\mathcal{M}$ and $\mathcal{A}$ being the action set included in the training set $\mathcal{D}_{train}$. In some occasions, the action set $\mathcal{A}$ is extremely large (e.g. 1000+ actions per task), so it may be infeasible to include all actions in the limited memory $\mathcal{M}$. We thus sample a subset of $h$ actions, $\mathcal{A}' \subseteq \mathcal{A}$, given the distribution of $P_{D}(\mathcal{A})$ in $\mathcal{D}_{train}$ where $p_{D}(a_{i})=\frac{n_{i}}{\sum_{{a_j} \in \mathcal{A}} n_{j}}$, with $n_i$ being the frequency of $a_{i}$ in $\mathcal{D}_{train}$. In that case, our method addresses the optimization problem over the actions in $\mathcal{A}'$. We solve the above problem by using an iterative updating algorithm, whose details can be found in
Appendix \ref{sec:sample}. The closest works~\cite{chrysakis2020online,kim2020imbalanced} maintain only the balanced label distribution in the memory while our work maintains the balanced memory w.r.t. both the LF and action distributions. 
\paragraph{Fast-Slow Continual Learning.}
Continual learning methods are expected to learn what the tasks have in common and in what the tasks differ. If there are some shared structures between tasks, it is possible to transfer knowledge from one task to another. Inspired by findings from cognitive neuroscience, the learning should be divided into slow learning of stationary aspects between tasks and fast learning of task-specific aspects~\cite{goyal2020inductiveBias}. This is an inductive bias that can be leveraged to obtain cross-task generalization in the space of all functions.

We implement this inductive bias by introducing a two-stages training algorithm. In the base model, action embeddings $\mathbf{c}_a$ (Eq. \eqref{eq:action_prob}) are task-specific, while the remaining parts of the model, which builds representations of utterances and action histories, are shared to capture common knowledge between tasks. Thus, in the \textit{fast-learning} stage, we update only the embeddings of unseen actions $\mathbf{c}^{(i)}_{a}$ with the cross-entropy loss, in the \textit{slow-learning} stage, we update all model parameters. Fast-learning helps parsers generalize to new tasks by giving the unseen actions good initialized embeddings and reduces the risk of forgetting prior knowledge by focusing on minimal changes of model parameters for task-specific patterns.

In the fast-learning stage, the unseen actions $A^{(k)}_{u}$ of the $k$-th task are obtained by excluding all historical actions from the action set of current task $\mathcal{T}^{(k)}$, namely $A^{(k)}_{u} = A^{(k)}\setminus A^{(1:k-1)}$, where $A^{(k)}$ denotes the action set of the $k$-th task. All actions are unseen in the first task, thus we update all action embeddings by having $A^{(0)}_{u} = A^{(0)}$. In the slow-learning stage, we differ updating parameters w.r.t. current task from updating parameters w.r.t. memories of previous tasks. For the former, the parameters $\theta_g$ shared across tasks are trained w.r.t all the data while the task-specific parameters $\theta^{(i)}_s$ are
trained only w.r.t. the data from task $\mathcal{T}^{(i)}$. For the latter, the task-specific parameters learned from the previous tasks are frozen to ensure they do not forget what is learned from previous tasks. 
\begin{algorithm}[t]
{\small
\SetKwData{Left}{left}\SetKwData{This}{this}\SetKwData{Up}{up}
\SetKwFunction{Union}{Union}\SetKwFunction{FindCompress}{FindCompress}
\SetKwInOut{Input}{Input}\SetKwInOut{Output}{Output}
\SetAlgoLined
\Input{Training set $\mathcal{D}^{(k)}_{train}$ of $k$-th task $\mathcal{T}^{(k)}$, memory data $\mathcal{M} = \{\mathcal{M}_1,...,\mathcal{M}_{k-1}\}$, the known action set $\mathcal{A}^{(1:k-1)}$ before learning $\mathcal{T}^{(k)}$}
Extract action set $\mathcal{A}^{(k)}$ from $\mathcal{D}^{(k)}_{train}$\\
Obtain unseen actions $\mathcal{A}^{(k)}_u$ by excluding $\mathcal{A}^{(1:k-1)}$ from $\mathcal{A}^{(k)}$\\
\textcolor{blue}{\# \textit{fast-learning on unseen action embeddings}}\\
\For{$i \gets 1$ to $epoch_1$} 
{
Update $\mathbf{c}^{(k)}_{a}$ with $\nabla_{\mathbf{c}^{(k)}_{a}}\mathcal{L}(\theta_{g},\theta^{(k)}_{s})$ on $\mathcal{D}^{(k)}_{train}$
}
\textcolor{blue}{\# \textit{slow-learning stage}}\\
    \For{$i \gets 1$ to $epoch_2$}{
    \textcolor{blue}{\# \textit{fine-tune all cross-task parameters and task-specific parameters of the current task}}\\
     Update $(\theta_{g},\theta^{(k)}_{s})$ with $\nabla_{(\theta_{g},\theta^{(k)}_{s})}\mathcal{L}(\theta_{g},\theta^{(k)}_{s})$ on $\mathcal{D}^{(k)}_{train}$\\
     \textcolor{blue}{\# \textit{replay training with task-specific parameters of the previous tasks frozen}}\\
    \For{$\mathcal{M}_i \in \mathcal{M}$}{
        Update $\theta_{g}$ with $\nabla_{\theta_{g}}\mathcal{L}(\theta_{g},\theta^{(i)}_{s})$ on $\mathcal{M}^{(i)}$\\
    }
    }
}
\caption{Fast-Slow Training for the $k$-th task }
\label{algo:adapter}
\end{algorithm}
More details can be found in Algo. \ref{algo:adapter}.
This training algorithm is closely related to Invariant Risk Minimization~\cite{arjovsky2019invariant}, which learns invariant structures across different training environments. However, in their work, they assume the same label space across environments and have access to all training environments at the same time. 

\paragraph{Loss} During training, we augment the \emr loss with the Elastic Weight Consolidation (\ewc) regularizer~\cite{kirkpatrick2017overcoming} to obtain the training loss $L_{CL} = L_{EMR} + \Omega_{EWC}$, where
$
    \Omega_{\ewc} = \lambda \sum^{N}_{j=1} F_{j}(\theta_{k,j}-\theta_{k-1,j})^{2}
$
, $N$ is the number of model parameters, $\theta_{k-1,j}$ is the model parameters learned until $\mathcal{T}^{(k-1)}$ and $F_{j} = \nabla^{2}\mathcal{L}(\theta_{k-1,j})$ w.r.t. the instances stored in $\mathcal{M}$. \ewc slows down the updates of parameters which are crucial to previous tasks according to the importance measure $F_{j}$.

\section{Experiments}
\label{sec:exp}
\paragraph{Datasets and Task Definitions.}
In this work, we consider three different scenarios: i) different tasks are in different domains and there are task-specific predicates and entities in LFs; 
ii) there are task-specific predicates in LF templates; iii) there are a significant number of task-specific entities in LFs. All tasks in the latter two are in the same domain. We select Overnight~\cite{wang2015building} and NLMapV2~\cite{lawrence2018improving} to simulate the proposed three continual learning scenarios, coined \overnight, \nlmapq and \nlmapc, respectively.


Overnight includes around 18,000 queries involving eight domains. The data in each domain includes 80\% training instances and 20\% test instances. Each domain is defined as a task. 

NLMapV2 includes 28,609 queries involving 79 cities and categorizes each query into one of 4 different question types and their sub-types. In the \nlmapq setting, we split NLMapV2 into 4 tasks with queries in different types. In the setting of \nlmapc, NLMapV2 is split into 8 tasks with queries of 10 or 9 distinct cities in each task. Each city includes a unique set of point of interest regions. In both \nlmapc and \nlmapq, each task is divided into 70\%/10\%/20\% of training/validation/test sets, respectively.

We attribute different distribution discrepancies between tasks to different definitions of tasks. Overall, distribution discrepancy between tasks on \overnight is the largest while the tasks in other two settings share relatively smaller distribution discrepancies because tasks of \nlmapq and \nlmapc are all in the same domain.

\paragraph{Baselines.}
Our proposed method is compared with 8 continual learning baselines. 
\textbf{FINE-TUNE} fine-tunes the model on the new tasks based on previous models. \textbf{EWC}~\cite{kirkpatrick2017overcoming} adds an L2 regularization to slow down the update of model parameters important to the historical tasks. \textbf{HAT}~\cite{serra2018overcoming} activates a different portion of parameters for each task with task-specific mask functions. \textbf{GEM}~\cite{lopez2017gradient} stores a small number of instances from previous tasks and uses the gradients of previous instances as the constraints on directions of gradients w.r.t. current instances. \textbf{EMR}~\cite{chaudhry2019tiny,wang2019sentence} trains the model on the data from the current task along with mini-batches of memory instances.  \textbf{EMAR}~\cite{han2020continual} is an extension of \emr using memory instances to construct prototypes of relation labels to prohibit the model from overfitting on the memory instances. \textbf{ARPER}~\cite{mi2020continual} adds an adaptive \ewc regularization on the \emr loss, where the memory instances are sampled with a unique sampling method called \prior. 
\textbf{ProtoParser}~\cite{li2021few} utilizes prototypical networks~\cite{snell2017prototypical} to improve the generalization ability of semantic parsers on the unseen actions in the new task. We customize it by training the \proto on the instances on current task as well as the memory instances. The \textbf{ORACLE} (All Tasks) setting trains the model on the data of all tasks combined, considered as an upper bound of continual learning.

\paragraph{Evaluation.} To evaluate the performance of continual semantic parsing, we report accuracy of exactly matched LFs as in~\cite{dong2018coarse}. We further adopt two common evaluation settings in continual learning. 
One setting measures the performance by averaging the accuracies of the parser on test sets of all seen tasks $\{\mathcal{D}_{test}^{(1)}$,...,$\mathcal{D}_{test}^{(k)}\}$ after training the model on
task $\mathcal{T}^{(k)}$, i.e. $\text{ACC}_{\text{avg}} = \frac{1}{k} \sum^{k}_{i=1} acc_{i,k}$~\cite{lopez2017gradient,wang2019sentence}.
The other one evaluates the test sets of all seen tasks combined after finishing model training on task $\mathcal{T}^{(k)}$, $
    \text{ACC}_{\text{whole}} = acc_{\mathcal{D}^{(1:k)}_{test}}
$~\cite{wang2019sentence,han2020continual}. For reproducibility, we include the detailed \textbf{implementation details} in Appendix \ref{sec:repro}.


\begin{table}[ht]
    \vspace{-2mm}
\centering
  \resizebox{0.95\textwidth}{!}{%
  \begin{tabular}{|c|c c|c c|cc|}
    \toprule
    \multirow{2}{*}{Methods} &
      \multicolumn{2}{c|}{\overnight} &
      \multicolumn{2}{c|}{\nlmapq} &
      \multicolumn{2}{c|}{\nlmapc} \\
      & W  & A & W  & A & W  & A \\
      \midrule
    Fine-tune & 14.40 & 14.37 &  60.22  & 55.53  & 49.29  & 48.11  \\
        \hline
    \ewc & 38.57 & 40.45 & 65.44 & 62.25  & 59.28& 57.56 \\
    \hatt & 15.45 & 15.71 & 64.69 & 60.88 & 53.30 & 52.41 \\
    \gem & 41.33  & 42.13 & 63.28 & 59.38  & 55.14 &54.37    \\
    \emr & 45.29 & 46.01 & 59.75 &55.59 & 58.36 & 56.95 \\
    \emar & 46.68 & 48.57 & 14.25 &12.89  & 51.79 & 50.93 \\
    \arper & 48.28 & 49.90  & 67.79 &64.73  & 62.62 &60.61\\
    \proto & 48.15 & 49.71  & 67.10 &63.73 & 62.58 &60.87 \\
    \hline
    \aemrt & 54.40 & 54.73 & 70.06 &67.77  & 62.96  & 61.59 \\
    \aemrt (+\ewc) & \textbf{59.02} & \textbf{58.04}  & \textbf{72.36} & \textbf{70.66} & \textbf{67.15}  & \textbf{64.89} \\
    \hline
    \oracle (All Tasks) & 63.14 & 62.76 & 73.20 & 71.65 & 69.48 & 67.18 \\
    \bottomrule
  \end{tabular}%
  }
    \caption{LF Exact Match Accuracy (\%) on two datasets with three settings after model learning on all tasks. “W” stands for the Whole performance $\text{ACC}_{\text{whole}}$, and “A” stands for the Average performance $\text{ACC}_{\text{avg}}$. All the results are statistically significant (p$<$0.005) compared with \aemrt(+\ewc) according to the Wilcoxon signed-rank test~\cite{woolson2007wilcoxon}. All experiments are run 10 times with different sequence orders and seeds. The memory size is 50 for all memory-based methods.\vspace{-2mm}}
  \label{tab:main}
  \vspace{-2mm}
\end{table}

\subsection{Results and Discussion}

\begin{figure}[t]
    \centering
    \includegraphics[width=0.9\textwidth]{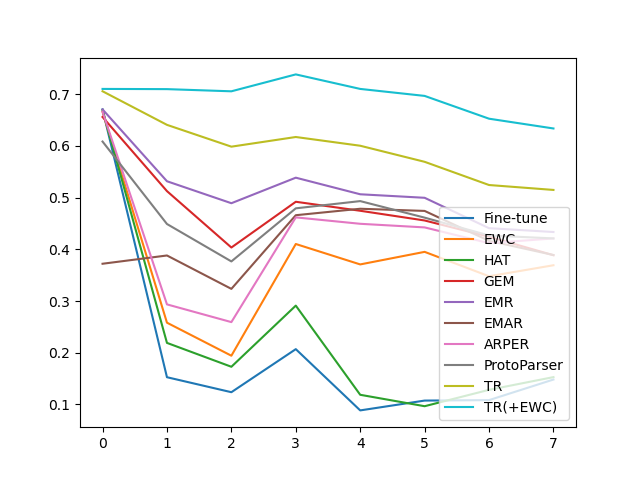}
    \caption{$\text{ACC}_{\text{whole}}$ till the seen tasks on \overnight after learning on each task sequentially. \vspace{-4mm}}
    \label{fig:acc_overnight}
        \vspace{-2mm}
\end{figure}

As shown in Table \ref{tab:main}, the base parser trained with our best setting, \aemrt(+\ewc), significantly outperforms all the other baselines (p$<$0.005) in terms of both $\text{ACC}_{\text{avg}}$ and $\text{ACC}_{\text{whole}}$. The performance of \aemrt(+\ewc) is, on average, only 3\% lower than the \oracle setting. Without \ewc, \aemrt still performs significantly better than all baselines except it is marginally better than \arper and \proto in the setting of \nlmapq. From Fig. \ref{fig:acc_overnight} we can see that our approaches are more stable than the other methods, and demonstrates less and slower forgetting than the baselines.


The dynamic architecture method, \hatt, performs worst on \overnight while achieves much better performance on \nlmapq and \nlmapc. Though the performance of the regularization method, \ewc, is steady across different settings, it ranks higher among other baselines on \nlmapc and \nlmapq than on \overnight. In contrast, the memory-based methods, \gem, and \emr, rank better on \overnight than on \nlmapq and \nlmapc. 

We conjecture that the overall performance of continual learning approaches varies significantly in different settings due to different distribution discrepancies as introduced in \textit{Datasets}. The general memory-based methods are better at handling catastrophic forgetting than the regularization-based and dynamic architecture methods, when the distribution discrepancies are large. However, those memory-based methods are less effective when the distribution discrepancies across tasks are small. 
Another weakness of memory-based methods is demonstrated by \emar, which achieves only 14.25\% of $\text{ACC}_{\text{whole}}$ on \nlmapq, despite it is the SOTA method on continual relation extraction. A close inspection shows that the instances in the memory are usually insufficient to include all actions when the number of actions is extremely large (i.e., more than 1000 actions per task in \nlmapq) while \emar relies on instances in memory to construct prototypes for each label. 
Furthermore, large training epochs for memory-based methods usually lead to severe catastrophic forgetting on the previous tasks, while the regularization method could largely alleviate this effect.


\arper and \proto are the two best baselines. Similar to \aemrt, \arper is a hybrid method combining \emr and \ewc, thus the joint benefits lead to consistent superior performance over the other baselines except \proto in all three settings. 
The generalization capability to unseen actions in new tasks also seems critical in continual semantic parsing. Merely combining \proto and \emr yields a new baseline, which performs surprisingly better than most existing continual learning baselines. From that perspective, the parser with \awu performs well in continual learning also because of its strength in generalizing to unseen actions.

\paragraph{Influence of Sampling Strategies.}
\begin{table}[ht]
    \vspace{-2mm}
\centering
  \resizebox{0.9\textwidth}{!}{%
  \begin{tabular}{|c|ccc|ccc|}
    \toprule
    \multirow{2}{*}{Methods} &
      \multicolumn{3}{c|}{\overnight} &
      \multicolumn{3}{c|}{\nlmapc} \\
    &10 &
      25 &
      50 & 
     10 &
      25 &
      50\\

      \midrule
    \random & 37.63 & 45.18 & 50.63 & 58.82 & 59.96 & 60.64 \\
    \fss & 38.64 & 47.08 & 52.63 & 58.94 & 59.67 & 60.89 \\
    \gss & 34.01 & 39.45 & 43.87 & 57.97 & 59.40&60.36  \\
    \prior & 37.60  & 44.84 & 50.14 &54.39  & 54.09 &53.43 \\
    \balance & 38.21 & 45.33 & 48.64 & 58.39 & 59.89 &61.19 \\
    \hline
    \lfs & 38.24 & 44.66 & 53.06 & 58.88 & 59.81 & 60.76 \\
    \blfst & 39.24 & 48.21 & 54.73 & 59.31 & 60.58 & 61.59 \\
    \bottomrule
  \end{tabular}%
  }
    \caption{$\text{ACC}_{\text{avg}}$ (\%) of \aemr with different sampling strategies and memory sizes 10, 25 and 50. \vspace{-4mm}}
  \label{tab:sample}
    \vspace{-2mm}
\end{table}
Table \ref{tab:sample} reports the evaluation results of \aemrt with different sampling strategies and sizes on \overnight and \nlmapc. The results on \nlmapq can be found in Appendix \ref{sec:sample_nlmapq}. \textbf{RANDOM} randomly samples instances from the train set of each task. \textbf{FSS}~\cite{aljundi2019gradient,wang2019sentence,mi2020continual}, \textbf{GSS}~\cite{aljundi2019gradient} and \textbf{LFS} partition the instances into clusters w.r.t. the spaces of utterance encoding features, instance gradients, and LFs, respectively, and then select the instances which are closest to the centroids. \textbf{PRIOR}~\cite{mi2020continual} selects instances that are most confident to the models and diversifies the entities in the utterances of memory. \textbf{BALANCE}~\cite{chrysakis2020online,kim2020imbalanced} balances the action distribution in a memory.

Overall, our sampling method consistently outperforms all other baselines on both \overnight and \nlmapc. On \overnight with memory size 50, the gap between \blfst and \gss is even up to 11\% and 2\% between \blfst and \fss, the best baseline. However, on \nlmapc, the performance differences across various sampling methods are smaller than those on \overnight. 
Similar observation applies to the influence of different sample sizes. 
We conclude that the smaller distribution discrepancy reduces the differences of sampling methods as well as the sample sizes in the memory-based methods.

\random performs steadily across different settings though it is usually in mediocre performance. \fss, \gss, and \prior are model-dependent sampling methods. The gradients and model confidence scores are not stable features for the sample selection algorithms. We inspect that the instances selected with \gss are significantly different even when model parameters are slightly disturbed. For the \prior, the semantic parsing model is usually confident to instances with similar LF templates. Diversifying entities do not necessarily lead to diversities of LF templates since the LFs with different entities may share similar templates. Therefore, \gss and \prior can only perform well in one setting. In contrast, the utterance encoding features are much more reliable. \fss can achieve the second-best performance among all methods. Either balancing action distribution (\balance) or selecting centroid LFs from LF clusters (\lfs) alone performs no better than \blfst, proving it is advantageous to select a instance in a cluster which balances the memory action distribution over directly using the centroid. 

\paragraph{Ablation Study of FSCL Training.}

\begin{table}[ht]
    \vspace{-2mm}
\centering
  \resizebox{0.9\textwidth}{!}{%
  \begin{tabular}{|c|c c|c c|cc|}
    \toprule
    \multirow{2}{*}{Methods} &
      \multicolumn{2}{c|}{\overnight} &
      \multicolumn{2}{c|}{\nlmapq} &
      \multicolumn{2}{c|}{\nlmapc} \\
      & W  & A & W  & A & W  & A \\
      \midrule
    
    \aemrt(+\ewc) & \textbf{59.02} & \textbf{58.04}  & \textbf{72.36}  & \textbf{70.66} & \textbf{67.15}  & \textbf{64.89} \\
    \quad - fast & 56.22 & 54.77  & 69.12 & 65.97 & 64.88 & 62.42\\
    -/+ fast/lwf & 55.80 & 54.54  & 69.45 & 66.37 & 65.14 & 62.70\\
    -/+ fast/emar & 56.93 & 56.05  & 69.43 & 66.42 &  64.89 & 62.51\\
    \hline
    \aemrt & 54.40 & 54.73 & 70.06  & 67.77  & 62.96  & 61.59 \\
    \quad - fast & 49.28 & 49.48  & 60.22 & 54.84 & 57.53 & 55.75\\
    -/+ fast/lwf & 47.47 & 47.77 & 58.96 & 53.54  & 55.24 & 56.89\\
    -/+ fast/emar & 49.63 & 48.74  & 64.98 & 61.09 & 56.89 & 55.24\\
    \bottomrule
  \end{tabular}%
  }
    \caption{The ablation study results of \awu.}
  \label{tab:abl_adap}
      \vspace{-3mm}
\end{table}
Table \ref{tab:abl_adap} shows the ablation study of 
\awu training by removing (-) or replacing (-/+) the corresponding component/step.

\begin{figure}[t]
    \vspace{-2mm}
    \centering
    \includegraphics[width=0.9\textwidth]{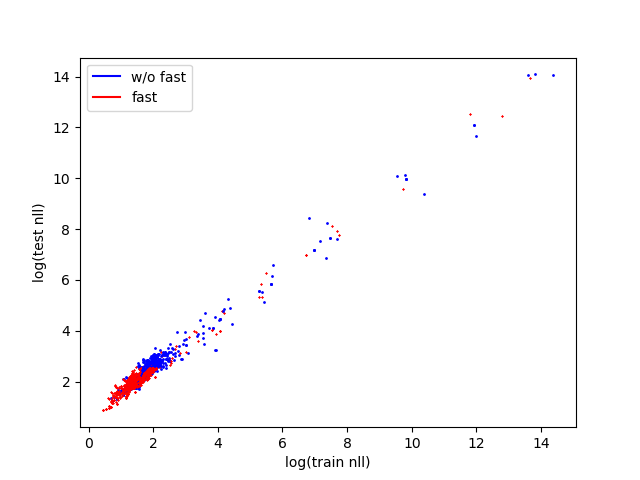}
    \caption{The training and test errors of the base parser with/without fast-learning on \overnight. \vspace{-4mm}}
    \label{fig:error_overnight}
\end{figure}

The fast-learning with action embeddings is the most critical step in \awu training. Removing it causes up to 13\% performance drop. To study this step in depth, we also replace our fast-learning with fine-tuning all task-specific parameters except in the first task, as done in LwF~\cite{li2017learning}, or fine tuning all parameters, as done in \emar~\cite{han2020continual}, in the fast-learning stage. The corresponding performance is no better than removing it in most cases. We also plot the training errors and test errors with or without this step in Fig. \ref{fig:error_overnight}. This step clearly leads to dramatically improvement of both generalization and optimization.

Another benefit of this fast-learning step is in the first task. We observe that a good optimization on the first task is crucial to the model learning on the following tasks. Our preliminary study shows that by applying the fast-learning only to the first task, the model can still keep the close-to-optimal performance. As shown in Fig. \ref{fig:acc_overnight}, our method with this fast-learning step is better optimized and generalized on the initial tasks than all the other baselines and largely alleviate the forgetting problem caused by learning on the second task. 

    \vspace{-2mm}

\paragraph{Influence of Pre-trained Language Models.}
We study the impact of pre-trained language models for semantic parsing in supervised learning and continual learning, respectively. In both settings, we evaluate the base parsers using BERT~\cite{devlin2019bert} as its embedding layer in two configurations: fine-tuning the parameters of BERT (BERT-finetune) and freezing BERT's parameters (BERT-fix). As in Tab. \ref{tab:bert}, BERT slightly improves the overall performance of the base parsers in supervised training (the \oracle setting) on \overnight. In contrast, in the continual learning setting, base parsers with the BERT embedding perform worse than the ones with the GLOVE embedding.
On \nlmapq, the accuracy of \finetune with GLOVE embedding is 30\% and 20\% higher than that with BERT's embedding updated and fixed, respectively. We conjecture the deeper neural models suffer more from the catastrophic forgetting. However, the average training speeds of parsers with BERT-fix and BERT-finetune are 5-10 times and 20-40 times respectively slower than those with GLOVE on each task. Overall, our method outperforms other SOTA continual learning methods except that \ewc with BERT-fix performs comparably with ours on \nlmapc. In contrast, the performance of \proto, the best baseline with GLOVE, is highly unstable on NLMap with BERT. 
\begin{table}[ht]
    \vspace{-2mm}
\centering
  \resizebox{0.95\textwidth}{!}{%
  \begin{tabular}{|c|c c|c c|cc|}
    \toprule
    \multirow{2}{*}{Methods} &
      \multicolumn{2}{c|}{\overnight} &
      \multicolumn{2}{c|}{\nlmapq} &
      \multicolumn{2}{c|}{\nlmapc} \\
      & W  & A & W  & A & W  & A \\
            \midrule
      \multicolumn{7}{|c|}{BERT-finetune} \\
        \midrule
    Fine-tune & 16.19 & 14.89 &  30.33  & 29.22  & 39.47  & 36.07 \\
        \hline
            \ewc  & 13.97 & 13.39 &  45.97 &  44.19   & 35.74& 34.17 \\
    \emr  & 51.74 &51.40 & 37.39 & 34.61 & 54.89 & 52.61 \\
            \proto & 47.74  & 46.99 &  15.10 & 13.03 & 34.56 & 32.27 \\
    \aemrt & 52.58 & 52.86 & 64.07 &61.35  & 58.38  & 56.35 \\

    \hline
    \oracle (All Tasks) & 65.40 & 64.39 & 72.70 & 71.60 & 67.51 & 68.80 \\
                \midrule
        \multicolumn{7}{|c|}{BERT-fix} \\
                  \midrule
        Fine-tune  & 14.10 & 12.09 &  40.45  & 36.06 & 49.15  & 46.48 \\
        \hline
            \ewc  & 17.69 & 18.56 & 54.20 & 51.32   &  64.10 & 61.67  \\
    \emr& 39.10 &39.03 & 42.46& 39.50 & 57.55 &55.42 \\
    \proto & 41.69 & 41.79 & 51.39 & 48.90 & 45.66 & 43.92 \\
    \aemrt & 47.14 & 48.12 & 59.41 & 55.84& 64.48  & 62.66 \\
    \hline
    \oracle (All Tasks) & 64.70 & 63.90 & 73.20 & 71.91 &69.80 & 67.57 \\
    \bottomrule
  \end{tabular}%
  }
    \caption{$\text{ACC}_{\text{whole}}$ and $\text{ACC}_{\text{avg}}$ (\%) of parsers using BERT by fine-tuning (Up) and fixing (Bottom) BERT's parameters.    \vspace{-3mm}}
  \label{tab:bert}
    \vspace{-3mm}
\end{table}

\section{Conclusion}
We conducted the first in-depth empirical study to investigate continual learning for semantic parsing. 
To cope with the catastrophic forgetting and facilitate knowledge transfer between tasks, we propose \tr, consisting of a sampling method specifically designed for semantic parsing and a two-stage training method implementing an inductive bias for continual learning. The resulted parser achieves superior performance over the existing baselines on three benchmark settings. The ablation studies also demonstrate why it is effective.

\section*{Acknowledgements}
This material is based on research sponsored by the ARC Future Fellowship FT190100039. The computational resources of this work are supported by the Multi-modal Australian Science Imaging and Visualisation Environment (MASSIVE). We appreciate the reviewers for their useful comments.

\bibliography{acl2021}
\bibliographystyle{acl_natbib}

\appendix
\BeforeBeginEnvironment{appendices}{\clearpage}
\begin{appendices}
\section{Reproducibility Checklist}
\label{sec:repro}
The hyper-parameters are cross-validated on the training set of \overnight and validated on the validation set of \nlmapq and \nlmapc. We train the semantic parser on each task with learning rate 0.0025, batch size 64 and for 10 epochs. The fast-learning training epochs is 5. We use the 200-dimensional GLOVE embeddings~\cite{pennington2014glove} to initialize the word embeddings for utterances. As different task orders influence the performance of the continual semantic parsing, all experiments are run on 10 different task orders with a different seed for each run. We report the average $\text{ACC}_{\text{avg}}$ and $\text{ACC}_{\text{whole}}$ of 10 runs. 
In addition, we use one GPU of Nvidia V100 to run all our experiments. The sizes of hidden states for LSTM and the action embeddings are 256 and 128, respectively. The default optimizer is Adam~\cite{kingma2014adam}. For our \dlfs method, we sample the subsets of the action sets with size 300 and 500 on \nlmapc and \nlmapq, respectively. The number of our model parameters is around 1.8 million. We grid-search the training epochs from \{10,50\}, the learning rate from \{0.001,0.0025\}. The coefficient for \ewc is selected from \{50000, 200000\}. Here we also provide the experiment results on the validation sets as in Table \ref{tab:valid}.
\begin{table}[ht]
\centering
  \resizebox{\textwidth}{!}{%
  \begin{tabular}{|c|c c|c c|cc|}
    \toprule
    \multirow{2}{*}{Methods} &
      \multicolumn{2}{c|}{\overnight} &
      \multicolumn{2}{c|}{\nlmapq} &
      \multicolumn{2}{c|}{\nlmapc} \\
      & W  & A & W  & A & W  & A \\
      \midrule
    \hline
    \aemrt & 52.38 & 53.30 & 68.78 &66.83  & 68.51  & 66.23 \\
    \aemrt (+\ewc) & 56.84 & 54.85  & 70.74 &69.21 & 74.84  & 71.19 \\
    \hline
    \bottomrule
  \end{tabular}%
  }
    \caption{Results on the validation sets.}
  \label{tab:valid}
    \vspace{-2mm}
\end{table}
\section{DLFS Algorithm}
\label{sec:sample}
We provide the detailed \blfs as in Algo. \ref{algo:sample}.
\begin{algorithm}[t]
{\small
\SetKwData{Left}{left}\SetKwData{This}{this}\SetKwData{Up}{up}
\SetKwFunction{Union}{Union}\SetKwFunction{FindCompress}{FindCompress}
\SetKwInOut{Input}{Input}\SetKwInOut{Output}{Output}
\SetAlgoLined
\Input{Training set $\mathcal{D}$, memory size $M$}
\Output{The memory $\mathcal{M}$}
     Partition $\mathcal{D}$ into $M$ clusters, denoted as $C$, with the similarity/distance metric\\
    Randomly sample the memory  $\mathcal{M}$ of size $M$ from $D$\\
    $H_{old} \gets -\inf$\\
     Compute the entropy $H_{new}$ of the action distribution $P_{\mathcal{M}}(\mathcal{A})$ in $\mathcal{M}$ as in Eq. \ref{eq:sample}\\
     \While {$H_{new} > H_{old}$}{
    \For{$i$-th cluster $c_i \in \mathcal{C}$}{
     Compute the entropy $H_{c}$ of $P_{\mathcal{M}}(\mathcal{A})$\\
    \For{$instance\ n \in c_i$}{
        Replicate $\mathcal{M}$ with $\mathcal{M}'$\\
        Replace the $i$-th instance $m_i$ in $\mathcal{M}'$ with $n$\\
        Compute the entropy $H'_{c}$ of $\mathcal{M}'$\\
        \If{$H'_{c} > H_{c}$}
        {
        $\mathcal{M} \gets \mathcal{M}'$\\
        $E_{c} \gets H'_{c}$
        }
    }
    $H_{old} \gets H_{new}$\\
    Compute the entropy $H_{new}$ of $P_{\mathcal{M}}(\mathcal{A})$\\
    }
    }
}
\caption{\blfst
}
\label{algo:sample}
\end{algorithm}
\section{Results of Different Sampling Methods}
\label{sec:sample_nlmapq}
Table \ref{tab:sample_nlmapq} shows the performance of \aemrt with different sampling strategies and different memory sizes on \nlmapq.
\begin{table}[ht]
\centering
  \resizebox{\textwidth}{!}{%
  \begin{tabular}{|c|ccc|}
    \toprule
    \multirow{2}{*}{Methods} &
      \multicolumn{3}{c|}{\nlmapq} \\
    &10 &
      25 &
      50 \\

      \midrule
    \random  & 65.63 & 66.20 & 67.21 \\
    \fss  & 66.51 & 67.15 & 67.67 \\
    \gss  & 65.89 & 66.60&67.07  \\
    \prior  &66.08  & 67.47 &67.42 \\
    \balance & 65.37 & 66.21 &66.99 \\
    \hline
    \lfs  & 66.08 &67.20 & 67.53 \\
    \blfst & 65.21 & 66.60 & 67.77 \\
    \bottomrule
  \end{tabular}%
  }
    \caption{$\text{ACC}_{\text{whole}}$ (\%) of \aemr with different sampling strategies and memory sizes 10, 25 and 50 on \nlmapq. }
  \label{tab:sample_nlmapq}
    \vspace{-2mm}
\end{table}

\section{Dynamic Action Representation} 
\label{app:dar}
To differentiate the learning of cross-task and task-specific aspects, we innovatively integrate a designated dynamic architecture into the base parser along with \dlfs and \awu for continual semantic parsing, coined \textbf{Dynamic Action Representation} (\dar). This method could also significantly mitigate the catastrophic forgetting and improve the forward transfer in the continual semantic parsing. Due to the limited space, we did not put it into the main paper. The details and analysis of this method are listed below.
\paragraph{Decoder of Base Parser.} The decoder of the base parser applies an LSTM to generate action sequences. At time $t$, the LSTM produces a hidden state $\mathbf{h}_t = \text{LSTM}(\mathbf{c}_{a_{t-1}}, \mathbf{h}_{t-1})$, where $\mathbf{c}_{a_{t-1}}$ is the embedding of the previous action $a_{t-1}$. We maintain an embedding for each action in the embedding table. As defined in~\newcite{luong2015effective}, we concatenate $\mathbf{h}_{t}$ with a context vector $\mathbf{o}_t$ to yield $\mathbf{s}_t$,

\vspace{-2mm}
\begin{small}
\begin{equation}
\label{eq:concate}
    \mathbf{s}_t = \tanh(\mathbf{W}_c [\mathbf{h}_t;\mathbf{o}_t])
\vspace{-1mm}    
\end{equation}
\end{small}
where $\mathbf{W}_c$ is a weight matrix and the context vector $\mathbf{c}_t$ is generated by the soft attention~\cite{luong2015effective},

\vspace{-5mm}
\begin{small}
\begin{equation}
\label{eq:attented_rep}
    \mathbf{o}_t = \sum_{i=1}^n \softmax(\mathbf{h}_t^{\intercal}\mathbf{E})\mathbf{e}_i
\end{equation}
\end{small}
The probability of an action $a_t$ is estimated by:

\vspace{-3mm}
\begin{small}
\begin{equation}
\label{eq:action_prob}
 P(a_t | \va_{<t},\vx) =  \frac{\exp(\mathbf{c}_{a_t}^{\intercal} \mathbf{s}_t)}{\sum_{a' \in \mathcal{A}_t}\exp(\mathbf{c}_{a'}^{\intercal} \mathbf{s}_t)}
\end{equation}
\end{small}
where $\mathcal{A}_t$ is the set of applicable actions at time $t$. In the following, the dense vectors $\mathbf{c}_{a}$ are referred to as \textit{action embeddings}.

\paragraph{Decoder of DAR.} The key idea of dynamic architectures is to add new parameters for new tasks and retain the previous parameters for old tasks~\cite{houlsby2019parameter,wang2020k,pfeiffer2020adapterfusion}. As a result, those methods can adapt to new tasks by using new parameters and still memorize the knowledge of previous tasks by keeping existing parameters. In semantic parsing, we differentiate task-specific actions $\mathcal{A}^{(k)}_s$, which generate task-specific predicates or entities, from cross-task actions, which are the remaining actions $\mathcal{A}_g$ associated with predicates appearing in more than one tasks. We model different actions using different action embeddings (Eq. \eqref{eq:action_prob}). But the key challenge lies in switching between task-specific and cross-task hidden representations.

To address the problem, given an output hidden state of LSTM, $\mathbf{h}_t = \text{LSTM}(\mathbf{c}_{a_{t-1}}, \mathbf{h}_{t - 1})$, we apply a task-specific adapter modules to transform the hidden state $\mathbf{h}_t \in \R^{d}$.
\begin{equation}
    \hat{\mathbf{h}}_t = g_i(\mathbf{h}_t) \phi_i(\mathbf{h}_t) + (1 - g_i(\mathbf{h}_t)) \mathbf{h}_t
\end{equation}
where $\phi_i(\cdot): \R^{d} \rightarrow \R^{d}$ is an adapter network and $g_i(\cdot): \R^{d} \rightarrow \R^{d} $ is a gating function for task $\mathcal{T}^{(i)}$. Here, we adopt the following modules for the adaptor network and the gating function,
\begin{align}
  \phi_i(\mathbf{h}_t) = \tanh(\mathbf{W}^{i}_{\phi}\textbf{h}_t)\\
  g_i(\mathbf{h}_t) = \text{sigmoid}(\mathbf{W}^{i}_{g}[\phi_i(\mathbf{h}_t);\mathbf{h}_t])
\end{align}
where parameters $\mathbf{W}^{i}_{\phi} \in \R^{d\times d} $ and $\mathbf{W}^{i}_{g} \in \R^{2d \times d}$ are task-specific. The number of parameters introduced per task is merely $O(3d^2)$, which is parameter-efficient. Therefore, the context vector of attention and the state to infer action probability in Eq. \ref{eq:concate} and \ref{eq:attented_rep} become:
\begin{align}
    \hat{\mathbf{o}}_t = \sum_{i=1}^n \softmax(\hat{\mathbf{h}}_t^{\intercal}\mathbf{E})\mathbf{e}_i\\
\hat{\mathbf{s}}_t = \tanh(\mathbf{W}_c [\hat{\mathbf{h}}_t;\hat{\mathbf{o}}_t])
\end{align}

\paragraph{Ablation Study of DAR}
\begin{table}[ht]
\centering
  \resizebox{0.9\textwidth}{!}{%
  \begin{tabular}{|c|c c|c c|cc|}
    \toprule
    \multirow{2}{*}{Methods} &
      \multicolumn{2}{c|}{\overnight} &
      \multicolumn{2}{c|}{\nlmapq} &
      \multicolumn{2}{c|}{\nlmapc} \\
      & W  & A & W  & A & W  & A \\
      \midrule
    
    \aemrt(+\ewc) & \textbf{59.02} & \textbf{58.04}  & \textbf{72.36}  & \textbf{70.66} & \textbf{67.15}  & \textbf{64.89} \\
    \quad - cross & 57.90 & 56.65 & 71.52 & 69.74  & 66.06 & 63.92\\
    \quad - specific & 55.32 & 54.64  & 71.47 & 69.81 & 65.87 & 63.73\\
    \hline
    \aemrt & 54.40 & 54.73 & 70.06  & 67.77  & 62.96  & 61.59 \\
    \quad - cross & 52.77 & 53.07  & 68.87 & 66.49 & 62.18 & 60.91\\
    \quad - specific & 51.44 & 52.84  & 69.18 & 66.88 & 61.26 & 59.93\\
    \bottomrule
  \end{tabular}%
  }
    \caption{The ablation study results of \dar.}
  \label{tab:abl_dar}
    \vspace{-2mm}
\end{table}
As shown in Tab. \ref{tab:abl_dar}, removing the task-specific representations (-specific) generally degrades the model performance by 1.5-3.5\% except on \nlmapq. Our further inspection shows that the proportion of task-specific actions in \nlmapq is only 1/20 while the ratios are 1/4 and 2/5 in \overnight and \nlmapc, respectively. Using either task-specific representations (-specific) or cross-task representations (-cross) alone cannot achieve the optimal performance.

\section{Accuracy Curve}
\begin{figure}[ht]
\centering
\begin{subfigure}[b]{\textwidth}
            \centering
    \includegraphics[width=\textwidth]{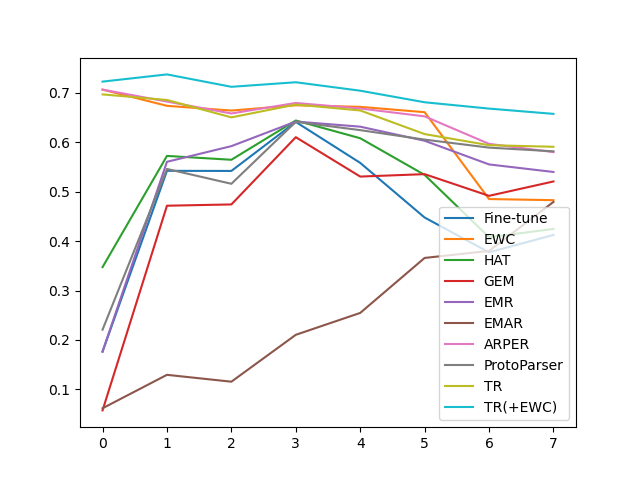}
\end{subfigure}

\begin{subfigure}[b]{\textwidth}
            \centering
    \includegraphics[width=\textwidth]{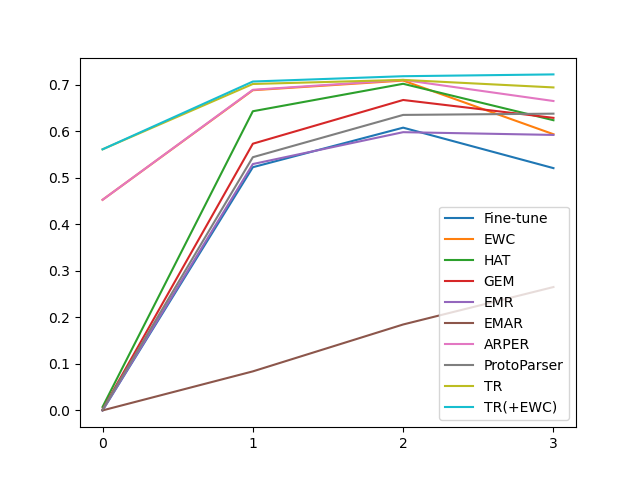}
\end{subfigure}
    \caption{$\text{ACC}_{\text{whole}}$ till the seen tasks on \nlmapc(Up) and \nlmapq(Bottom) after learning on each task sequentially.}
    \label{fig:acc_nlmap}
\end{figure}
\begin{table}[ht]
\centering
  \resizebox{\textwidth}{!}{%
  \begin{tabular}{|c|ccc|}
    \toprule
    Methods & \overnight & \nlmapq
       &
      \nlmapc \\

      \midrule
    Fine-tune & 54.19 &  360.09 & 110.12  \\
        \hline
    \ewc &  71.66 &  541.20 & 223.02  \\
    \hatt &  53.85 &  212.63 & 128.67  \\
    \gem &  94.67 &  389.04 & 259.80     \\
    \emr &  102.97 &  399.05 & 214.95  \\
    \emar &  139.44 &  402.64 & 240.64  \\
    \arper & 160.05 &  901.33 & 654.51 \\
    \proto & 148.04 &  490.95 & 304.45  \\
    \hline
    \aemrt &  117.63 &  549.22 & 275.09  \\
    \aemrt (+\ewc) & 124.49 &  540.68 & 282.87  \\
    \hline
    \oracle (All Tasks) &  712.43 &  1876.60 & 1531.05  \\
    \bottomrule
  \end{tabular}%
  }
    \caption{The average training time (seconds) of each continual learning method on one task. The training time of the \oracle setting is reported with training on all tasks. All the methods are running on a server with one Nvidia V100 and four cores of Intel i5 5400.}
  \label{tab:walltime}
    \vspace{-2mm}
\end{table}
\begin{figure}[ht]
    \centering
    \begin{subfigure}[b]{\textwidth}
        \centering
    \includegraphics[width=\textwidth]{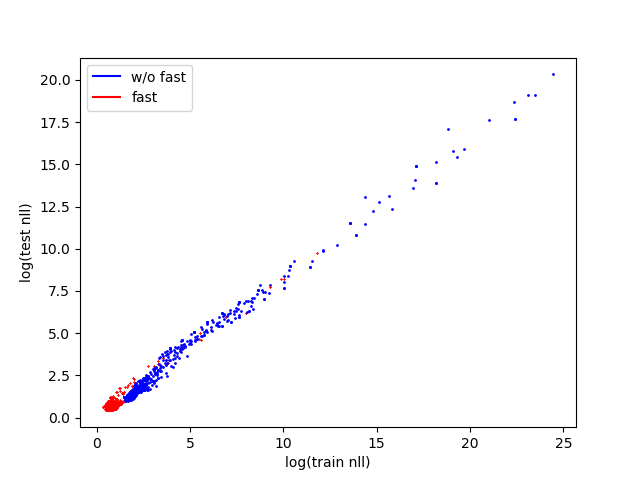}
    \end{subfigure}
    \begin{subfigure}[b]{\textwidth}
                \centering
    \includegraphics[width=\textwidth]{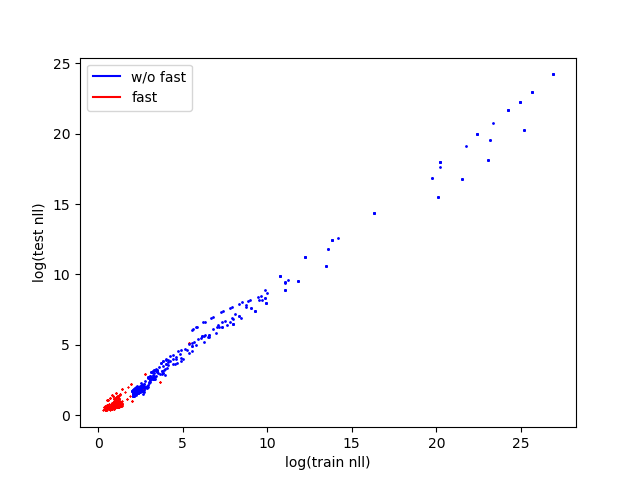}

                    \end{subfigure}
    \caption{The training and test error points of semantic parsing models with/without fast-learning on \nlmapc (Up) and \nlmapq (Bottom).}
    \label{fig:error_nlmap}
\end{figure}
\begin{figure}[ht]
    \centering
        \begin{subfigure}[b]{\textwidth}
    \centering
    \includegraphics[width=\textwidth]{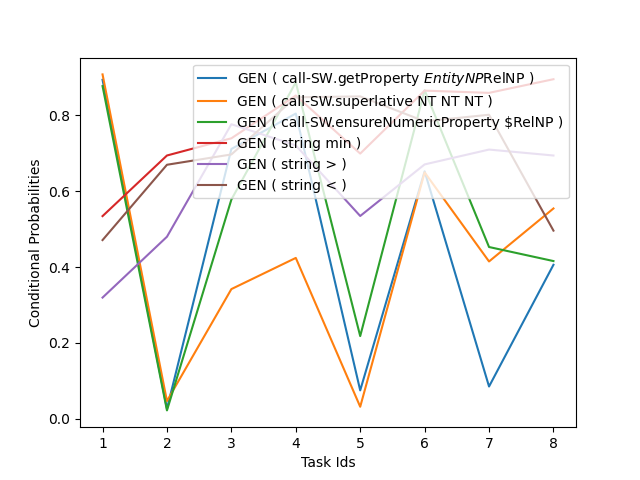}
    \end{subfigure}
    
    \begin{subfigure}[b]{\textwidth}
            \centering
    \includegraphics[width=\textwidth]{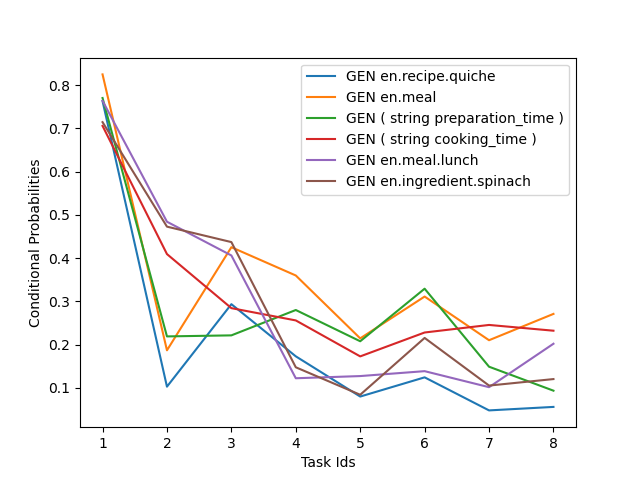}
    \end{subfigure}
    \caption{The conditional probabilities $P(a_t | \va_{<t},\vx)$ of the representative cross-task actions (Up) and task-specific actions (Bottom) from evaluation on the initial task after parser being trained on each task on \overnight sequentially.}
    \label{fig:cond_prob_overnight}
\end{figure}
Figs. \ref{fig:acc_nlmap} depicts the performance curve of semantic parsers till the seen tasks on \nlmapc and \nlmapq after learning on each task sequentially.

The base parsers are the same for all training methods in comparison. However, the training methods are not exactly the same. For example, \proto and \emar use meta-learning methods to train the parser. \hatt manipulates parameter gradients during training and uses adapter layers to modify the weights of model parameters on different tasks. \arper and \ewc use regularization during continual training. Different training methods cause the baselines to obtain different results on the initial and subsequent tasks.

In the first task, Fast-Slow Continual Learning (\awu) differs from the traditional supervised training by updating all action embeddings first, followed by updating all model parameters. From Fig. \ref{fig:acc_overnight} and Fig. \ref{fig:acc_nlmap}, we can tell \awu leads to a significant performance gain over the baselines in the first task. In this way, our parser trained with \awu lays a better foundation than the baselines for learning future tasks in terms of both the forward and backward transfer. For the new tasks, the fast-learning step of \awu leads to minimal changes of model parameters for task-specific patterns. In contrast, the baselines modify the majority of model parameters for each new task, hence easily lead to catastrophic forgetting. As a result, our model with \awu could achieve better performance than all baselines on both all tasks and only the initial task as in Fig. \ref{fig:acc_overnight} and Fig. \ref{fig:acc_nlmap}.

\section{Training Time Analysis}
The average training times of different continual learning models on each task of \overnight, \nlmapc, and \nlmapq are depicted in Tab. \ref{tab:walltime}. On average, the training time of Fine-tune is 13, 5, and 14 times faster than training the parser from scratch on the tasks of \overnight, \nlmapc, and \nlmapq, respectively. In general, the training times of memory-based methods are longer than regularization and dynamic architecture methods due to the replay training. Since our method, \tr, is a memory-based method, its training time is comparable to the other memory-based methods such as \gem, \emr and \emar. In addition, \ewc slowers the convergence speed of the parser on \nlmapc, and \nlmapq, thus increases the training time of parsers on each task to achieve their optimal performance. Therefore, the hybrid method, \arper, that utilizes both \emr and \ewc takes the longest training time among all continual learning methods. However, our \awu could speed up the convergence of the base parser even with \ewc; thus, the training time of \tr (+EWC) is much less than the one of \arper.

\vfill
\section{Training and Test Error Plots}
Fig. \ref{fig:error_nlmap} provides the training and test error points of semantic parsers on \nlmapc and \nlmapq, respectively. As we can see, same as on \overnight, the base parser with this fast-learning step is better optimized than without this step on \nlmapc and \nlmapq.


\vfill
\section{Forgetting Analysis on Actions}
Following \ref{sec:challenges}, Fig. \ref{fig:cond_prob_overnight} depicts the conditional probabilities, $P(a_t | \va_{<t},\vx)$, of cross-task and task-specific actions, respectively, predicted by the base parser fine-tuned sequentially on each task. Overall, task-specific actions are more likely to be forgotten than cross-task actions while learning parsers on the new tasks. Due to the rehearsal training of the cross-task actions in the future tasks, the prediction performance over cross-task actions fluctuates on different tasks.




\end{appendices}

\end{document}